\documentclass[10pt,twocolumn,letterpaper]{article}

\usepackage[pagenumbers]{cvpr} 

\usepackage{graphicx}
\usepackage{amsmath}
\usepackage{amssymb}
\usepackage{booktabs}
\usepackage{bm}
\usepackage{multirow}
\usepackage{multicol}

\usepackage[pagebackref,breaklinks,colorlinks]{hyperref}

\usepackage[linesnumbered,ruled,vlined]{algorithm2e}
\usepackage{algorithmic}
\usepackage{ragged2e}

\newcommand{\btheta}{\bm{\theta}}

\newcommand{\mt}{\mathcal{T}}
\newcommand{\ms}{\mathcal{S}}

\newcommand{\yq}[1]{{\color{black}{#1}}}

\DeclareMathOperator*{\argmin}{arg\,min}

\usepackage[capitalize]{cleveref}
\crefname{section}{Sec.}{Secs.}
\Crefname{section}{Section}{Sections}
\Crefname{table}{Table}{Tables}
\crefname{table}{Tab.}{Tabs.}

\def\cvprPaperID{732} 
\def\confName{CVPR}
\def\confYear{2022}

\begin{document}

\title{Improved Distribution Matching for Dataset Condensation}

\author{Ganlong Zhao$^{1,2}$ \quad Guanbin Li$^{1,3}$\thanks{Corresponding authors are Guanbin Li and Yizhou Yu.} \quad Yipeng Qin$^{4}$ \quad Yizhou Yu$^{2*}$\\
$^1${Sun Yat-sen University} \quad $^2$The University of Hong Kong\\ 
$^3${Research Institute, Sun Yat-sen University, Shenzhen} \quad $^4${Cardiff University} \\
{\tt\small  zhaogl@connect.hku.hk, liguanbin@mail.sysu.edu.cn, qiny16@cardiff.ac.uk, yizhouy@acm.org}
}

\maketitle

\begin{abstract}
   Dataset Condensation aims to condense a large dataset into a smaller one while maintaining its ability to train a well-performing model, thus reducing the storage cost and training effort in deep learning applications. However, conventional dataset condensation methods are optimization-oriented and condense the dataset by performing gradient or parameter matching during model optimization, which is computationally intensive even on small datasets and models. In this paper, we propose a novel dataset condensation method based on distribution matching, which is more efficient and promising. 
   Specifically, we identify two important shortcomings of naive distribution matching (\ie, imbalanced feature numbers and unvalidated embeddings for distance computation) and address them with three novel techniques (\ie, partitioning and expansion augmentation, efficient and enriched model sampling, and class-aware distribution regularization).
   Our simple yet effective method outperforms most previous optimization-oriented methods with much fewer computational resources, thereby scaling data condensation to larger datasets and models.
   Extensive experiments demonstrate the effectiveness of our method. Codes are available at \url{https://github.com/uitrbn/IDM}
\end{abstract}

\section{Introduction}
\label{sec:intro}

Deep learning\cite{lecun1998gradient,krizhevsky2017imagenet,NMI21} is notoriously data-hungry, which poses challenges for both its training and data storage.
To improve data storage efficiency, Dataset Condensation (DC)\cite{wang2018dataset,zhao2021dataset} aims to condense large datasets into smaller ones while retaining their validity for model training.
Unlike traditional coreset selection methods~\cite{welling2009herding,wolf2011facility,sener2018active,toneva2018empirical}, such dataset condensation is often achieved through image synthesis and yields better performance.
If properly condensed, the resulting datasets not only consume less storage space, but can also benefit various downstream tasks such as network architecture search and continual learning by reducing their computational costs. 

\begin{figure}[t]
  \centering
   \includegraphics[width=0.95\linewidth]{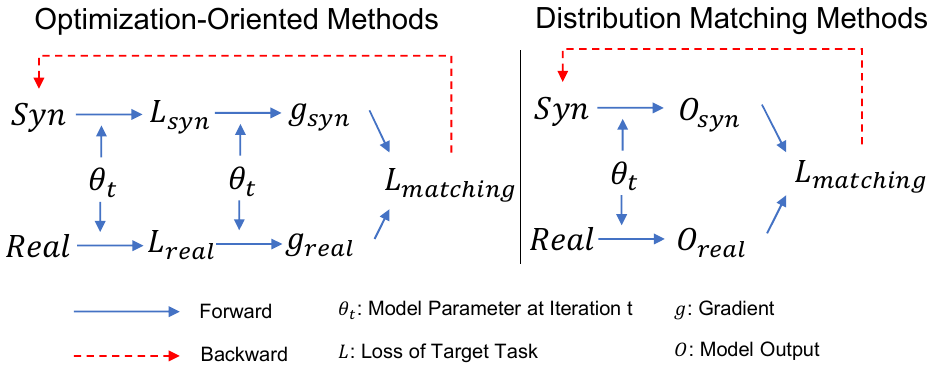}

\caption{Illustration of optimization-oriented methods and distribution matching methods. $L$: classification loss; $g$: gradient; $O$: output of models; $L_{matching}$: the matching loss for condensation.}
\vspace*{-6mm}

   \label{fig:figure1}
\end{figure}

The pioneering DC approach\cite{zhao2021dataset,zhao2021dataset} guarantees the validity of a condensed dataset by imposing a strong assumption that a model trained on it should be identical to that trained on the real dataset. However, naive matching between these converged models can be too challenging due to their large parameter space and long optimization towards convergence.
To this end, they impose an even stronger assumption that the two models should share an identical or similar optimization path, which can be achieved by matching either their gradients\cite{zhao2021dataset,zhao2021DSA} or their intermediate model parameters\cite{cazenavette2022dataset} during training.
We therefore refer to them as {\it optimization-oriented} methods.

However, despite their success, the unique characteristics of optimization-oriented DC methods imply that they inevitably suffer from high computational costs and thus scale poorly to large datasets and models.
Specifically, they all involve the optimization of models (either randomly-initialized~\cite{zhao2021dataset,zhao2021DSA} or initialized with parameters of pretrained models~\cite{cazenavette2022dataset}) against the condensed dataset, and thus rely on a nested loop that optimizes the condensed dataset and model parameters in turn.
Note that the use of pretrained models require additional computation and storage space~\cite{cazenavette2022dataset}.
As a result, existing optimization-oriented DC methods are only applicable to ``toy'' networks (\eg, three-layer convolutional networks) and small datasets (\eg, CIFAR10, CIFAR100).
Whether they can be scaled to real-world scenarios is still an open question.

To scale DC to large models and datasets, distribution matching (DM)~\cite{zhao2021DM} proposes to match the output feature distributions of the real and condensed datasets extracted by randomly-initialized models.
This stems from the fact that the validity of a condensed dataset can also be guaranteed if it produces the same feature distribution as the real dataset.
Since DM does not involve the optimization of models against the condensed dataset, it avoids the expensive nested loops in optimization-oriented methods, and is thus highly efficient and scalable.
However, despite being promising, experimental results show that its performance still lags behind that of the state-of-the-art optimization-oriented methods.

In this paper, we perform an in-depth analysis on DM's unsatisfactory performance and propose a set of remedies that can significantly improve it, namely improved distribution matching (IDM).
Specifically, we analyzed the output feature distributions of DM and observed that although their means match, the features of the condensed dataset scatter around, causing severe class misalignment problems, which accounts for its impaired performance.
We ascribe such scattered features to two shortcomings of DM as follows:

\noindent
i) DM suffers from the imbalanced number of features.
Intuitively, DM uses the features of {\it small} condensed datasets to match those of {\it large} real datasets, which is inherently intractable.
Addressing this shortcoming, we propose {\it Partitioning and Expansion augmentation}, which augments the condensed dataset by evenly splitting each image into $l \times l$ parts and expanding each part to the size of the original image, resulting in $l^2$ features per image and a better match to the features of the real dataset.

\noindent
ii) Randomly initialized models are not valid embedding functions for the Maximum Mean Discrepancy (MMD)~\cite{gretton2012kernel} estimation used in DM. 
Specifically, DM justifies the validity of randomly-initialized models by their intrinsic classification power observed in tasks such as deep clustering~\cite{caron2018deep,saxe2011random,cao2018review,amid2022learning}. However, we believe that this does not apply to DM as randomly-initialized models do not satisfy the requirement of embedding functions used in MMD and makes it an invalid measure of distribution distance.
Since it is too challenging to design neural network based embedding functions that are valid for MMD, we propose two simple yet effective remedies:
1) {\it Efficient and enriched model sampling}.
We enrich the embedding functions in MMD with semi-trained models as additional feature extractors, and develop a memory-efficient model queue to facilitate their sampling.
2) {\it Class-aware distribution regularization}. We explicitly regularize the feature distributions of condensed datasets to further alleviate class misalignment.

\noindent
These three novel techniques together help to extract better feature distributions for DM.
Our contributions include: 
\begin{itemize}
    \item We deeply analyze the shortcomings of the Distributed Matching~\cite{zhao2021DM} algorithm and reveal that the root of its impaired performance lies in the problem of class misalignment.
    \item We propose improved distribution matching (IDM), consisting of three novel techniques that address the shortcomings of DM and help to learn better feature distributions. 
    \item Experimental results show that our IDM achieves significant improvement over DM and surpasses the performance of most optimization-oriented methods.
    \item We show that our IDM method is highly efficient and scalable, and can be applied to large datasets such as ImageNet Subset\cite{deng2009imagenet,tian2020contrastive}.
\end{itemize}

\section{Related Works}

\noindent \textbf{Dataset Condensation.} Dataset condensation aims to condense large datasets to smaller ones while preserving the information to train models. 
It can benefit various applications including continual learning\cite{zhao2021dataset, zhao2021DM, zhao2021DSA}, efficient neural architecture search\cite{zhao2021dataset, zhao2021DSA, zhao2021DM}, federated learning\cite{goetz2020federated,sucholutsky2020secdd,zhou2020distilled} and privacy-preserving ML\cite{li2020soft,dong2022privacy}.
Data Distillation (DD)\cite{wang2018dataset} pioneered this topic by maximizing the accuracy of models trained by the condensed set with meta-learning techniques\cite{nichol2018first}.
Later methods significantly outperformed DD by introducing more advanced techniques, such as soft-label\cite{bohdal2020flexible,sucholutsky2021soft}, gradient matching\cite{zhao2021dataset}, augmentation\cite{zhao2021DSA}, infinite kernel-limit\cite{nguyen2020dataset, nguyen2021dataset}, long-range parameter matching\cite{cazenavette2022dataset}, data parameterization\cite{pmlr-v162-kim22c}, contrastive signal\cite{lee2022dataset} and feature alignment\cite{wang2022cafe}. 
Despite their success, most of the best-performing methods rely on bi-level optimizations involving second-order derivatives and thus require intensive computation resources.
Recently, some researchers proposed to use subnet optimization to reduce the computation cost\cite{zhou2022dataset}, but it still cannot condense large datasets due to its intensive kernel computation. 

In contrast, Distribution Matching (DM)\cite{zhao2021DM} discards the bi-level optimization and condenses datasets by matching the feature distributions of the real and condensed sets. 
This saves a lot of memory and computation, allowing DM to condense large datasets.
However, such high efficiency comes at the cost of inferior performance, which hinders the further application of DM. 
In this work, we address two important shortcomings of DM, thereby scaling it to larger datasets and models without sacrificing performance.

Some other methods adopted a generative modeling approach for data condensation\cite{masarczyk2020reducing,such2020generative,zhao2022synthesizing}.
However, we do not compare with them due to the different settings.

\vspace{2mm}
\noindent \textbf{Coreset Selection.} 
Coreset selection methods~\cite{agarwal2004approximating,har2004coresets,feldman2020turning,chen2012super,wei2015submodularity,feldman2020introduction,feldman2011scalable,feldman2007ptas,yoon2021online}, which have been widely used in continual learning~\cite{rebuffi2017icarl,toneva2018empirical,castro2018end,aljundi2019gradient} and active learning~\cite{sener2017active}, first describe a criterion to measure the representativeness of samples, which is then used to identify and cluster the coreset.
Examples of these criteria include compactness\cite{rebuffi2017icarl}, diversity~\cite{sener2017active,aljundi2019gradient}, and forgetfulness\cite{toneva2018empirical}. 
However, these heuristic criteria are irrelevant to target tasks and can not guarantee the optimal solution. 
In addition, the performance of coreset selection is restricted by the quality of the original images.

\begin{figure*}[t]
  \centering
   \includegraphics[width=0.85\linewidth]{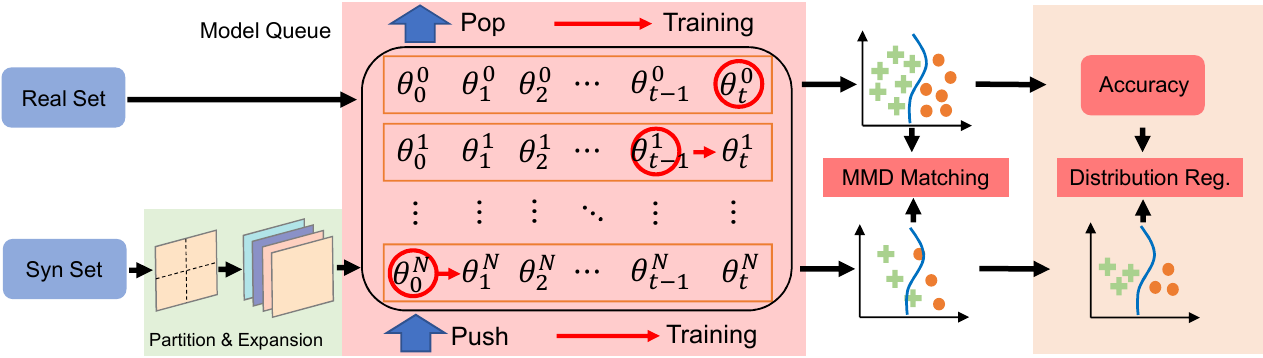}

   \caption{The overall framework of our method. Images in the synthetic set are first partitioned and expanded (Sec.~\ref{sec:augmentation}) and then fed with those of the real set into the model sampled from our efficient and enriched model queue (Sec.~\ref{sec:efficient_enriched_sampling}). The matching of output distributions is regularized with our class-aware distribution regularization (Sec.~\ref{sec:class_aware_distribution_regularization}). Push, Pop, and Training: model queue operations. }
   \vspace{-3mm}
   \label{fig:framework}
\end{figure*}

\section{Problem Definition}

\noindent 
\textbf{Dataset Condensation.} 
Given a large training set $\mathcal{T} = \{(\bm{x}_1, y_1), ..., (\bm{x}_{|\mathcal{T}|}, y_{|\mathcal{T}|})\}$ containing $|\mathcal{T}|$ images and their labels, dataset condensation aims to synthesize a much smaller set $\mathcal{S} = \{(\bm{s}_1, y^s_1), ... ,(\bm{s}_{|\mathcal{S}|}, y^s_{|\mathcal{S}|})\}$, $|\mathcal{S}| \ll |\mathcal{T}|$ such that $\mathcal{S}$ has the same or similar power as $\mathcal{T}$ in terms of model training. 
Let $\bm{x}$ be a sample from the real data distribution $P_D$ with label $y$, $\phi_{\theta^\mathcal{T}}$ and $\phi_{\theta^\mathcal{S}}$ be two variants of the same model $\phi$ with parameters $\theta$ trained on $\mathcal{T}$ and $\mathcal{S}$, respectively, $\ell$ be the loss function (cross-entropy loss), we have:
\begin{equation}
    \mathcal{S^*} =\argmin_\mathcal{S} \mathbb{E}_{\bm{x} \sim P_D} || \ell(\phi_{\theta^\mathcal{T}}(\bm{x}), y) -
     \ell(\phi_{\theta^\mathcal{S}}(\bm{x}), y) ||,
\label{eq:definition}
\end{equation}

\noindent
\textbf{Distribution Matching.}
Previous methods either solved Eq.~\ref{eq:definition} as a nested optimization problem directly \cite{wang2018dataset} or converted it to an gradient/parameter matching problem \cite{zhao2021dataset}, which are all computationally intensive and scale poorly to large datasets.
To overcome this computational barrier, Distribution Matching (DM)~\cite{zhao2021DM} proposed to match the feature distributions $\phi_{\theta}(\bm{x}_i)$ and $\phi_{\theta}(\bm{s}_j)$ for $\mathcal{T}$ and $\mathcal{S}$, respectively. 
Taking maximum mean discrepancy (MMD)~\cite{gretton2012kernel} as the distance measure, it can be formulated as:
\begin{equation}
\begin{aligned}
    \mathcal{S^*} &= \argmin_\mathcal{S} \mathcal{L}_{DM_{\theta \sim P_{\theta_0}}} \\
    &= \argmin_\mathcal{S} \mathbb{E}_{\theta \sim P_{\theta_0}} || \frac{1}{|\mathcal{T}|} \sum_{i=1}^{|\mathcal{T}|} \phi_{\theta}(\bm{x}_i) - \frac{1}{|\mathcal{S}|} \sum_{j=1}^{|\mathcal{S}|} \phi_{\theta}(\bm{s}_j)||^2,
\end{aligned}
    \label{eq:dm_objective}
\end{equation}
where $P_{\theta_0}$ denotes the distribution of randomly initialized network parameters.
Note that Eq.~\ref{eq:dm_objective} relies on the random initialization of $\theta \sim P_{\theta_0}$ and avoids its training, thereby reducing the computational cost.

\section{Methodology}

Recognizing the great potential of Distribution Matching (DM) \cite{zhao2021DM} for data condensation, we use it as the starting point for our research.
Specifically, we aim to address two of its shortcomings (SCs) described below.

\vspace{2mm}
\noindent {\bf SC1: Imbalanced Number of Features between $\mathcal{T}$ and $\mathcal{S}$.} DM aims to match the feature distribution of $\mathcal{T}$ with that of $\mathcal{S}$ with $|\mathcal{S}| \ll |\mathcal{T}|$. However, this is inherently intractable given such a small $|\mathcal{S}|$.
Specifically, in naive DM, one feature is extracted for each image in either $\mathcal{S}$ or $\mathcal{T}$.
Thus, the above aim indicates that DM works well if and only if the large number of features (\eg, thousands) extracted from $\mathcal{T}$ can be well approximated by a much smaller number of features (\eg, ten) extracted from $\mathcal{S}$. This would only be the case when $\mathcal{T}$ is highly redundant, which is not true for most real-world scenarios.

\vspace{2mm}
\noindent {\bf SC2: Unvalidated Embeddings in MMD Computation.}
DM used maximum mean discrepancy (MMD) to measure the distance between the feature distributions of real and synthetic datasets (Eq.~\ref{eq:dm_objective}).
However, instead of carefully designing the mapping functions, DM proposed to use randomly initialized networks $\phi_\theta$ to get various embeddings for MMD estimation directly, whose validity has not been verified.
Specifically, DM claimed that randomly initialized models, \ie, $\theta \sim P_{\theta_0}$, are sufficient for feature extraction and comparable to trained models. 
However, randomly initialized models are inadequate as their parameters are sampled from a simple pre-defined distribution, which has a limited number of patterns and occupies only a small fraction of the hypothesis space. Besides, the equivalence between gradient and distribution matching\cite{zhao2021DM} reveals the necessity of distribution matching throughout the entire optimization procedure, not just the initialization stage.

\subsection{Partitioning and Expansion Augmentation}
\label{sec:augmentation}

Addressing {\bf SC1}, we propose a simple yet effective technique, namely partitioning and expansion augmentation, to increase the number of features extracted from $\mathcal{S}$ without increasing its size $|\mathcal{S}|$.
Specifically, for each image $s_i \in \mathcal{S}$, we first partition it into $l\times l$ equal pieces, and then expand each piece to the size of $s_i$ using differentiable augmentation~\cite{zhao2021DSA}:
\begin{equation}
    s_i^1, s_i^2, ... s_i^{l \times l} = \text{Expand}(\text{Partition}(s_i, l)).
\end{equation}
In this way, we increase the number of features extracted from $\mathcal{S}$ from $|\mathcal{S}|$ to $l^2|\mathcal{S}|$ without increasing its size, facilitating DM by alleviating the imbalance in feature numbers identified in SC1.

\vspace{2mm}
\noindent \textbf{Remark.}
The rationale of this augmentation stems from our observation that the potential of synthetic images in $\mathcal{S}$ was underutilized. Specifically, it is well known that many fine details of an image are discarded during feature extraction and contribute little to the final results.
Based on this observation, we discard the fine details with our partitioned pieces before feature extraction but retain their power in semantic representation of an image, which makes better use of the synthetic images.
We observed some concurrent works~\cite{pmlr-v162-kim22c} with similar ideas. However, they employed an optimization-oriented approach and did not recognize the unique benefits of this augmentation to DM.

\subsection{Efficient and Enriched Model Sampling}
\label{sec:efficient_enriched_sampling}

Addressing {\bf SC2}, we propose to extend the $\theta \sim P_{\theta_0}$ used in Eq.~\ref{eq:dm_objective} to $\theta \sim P_{\theta(T)}$ where $P_\theta(T) = P_{\theta_0} \cup P_{\theta_1} \cup ... \cup P_{\theta_T}$, $P_{\theta_T}$ denotes the parameter distribution of models initialized with $P_{\theta_0}$ and trained for $T$ iterations.
Intuitively, this extension {\it enriched} the sampling of $\theta$ with an additional dimension ``training iterations'', allowing for a significant diversification of the parameter distribution so that more informative features can be extracted for data condensation.

In practice, a naive way to construct $P_\theta(T)$ is to initialize models with samples from $P_{\theta_0}$ and record their intermediate parameters during training as estimations of $P_{\theta_t}$, ($0 \leq t \leq T$).
However, this may not be feasible as it suffers from the trade-off between $T$ and the computational resources required to pre-train and store the model parameter samples: the larger $T$, the more diverse $P_\theta(T)$, but also the more computation costs. 
To mitigate this trade-off, a straightforward idea is to identify and remove the redundancies in $P_\theta(T)$.
Nevertheless, according to previous studies~\cite{zhao2021dataset}, the intermediate model parameters $\theta_t$ is highly dependent on its initial value $\theta_0$ and different from each other, which implies that the elements in $P_\theta(T)$ are mostly necessary and rarely redundant.
Addressing this challenge, we propose a novel data structure
to facilitate {\it efficient} model sampling as follows.

\vspace{2mm}
\noindent \textbf{Memory-efficient Model Queue.}
As an effective estimation of $P_\theta(T)$, our memory-efficient model queue $Q$ has four operations: 
\begin{itemize}
    \item $\mathrm{Push}$. Push a newly initialized model $\phi_\theta$ to $Q$, where $\theta \sim P_{\theta_0}$. 
    \item $\mathrm{Pop}$. Pop a model from $Q$ if $Q$'s size exceeds $N_{\mathrm{max}}$.
    \item $\mathrm{Train}$. Randomly fetch a model from $Q$ and train it using real data for $K$ iterations. 
    \item $\mathrm{Sample}$. Randomly sample a model from $Q$ for use.
\end{itemize}
Among them, $\mathrm{Sample}$ implements the sampling $\theta \sim P_\theta(T)$ for data condensation and the rest three are used to maintain $Q$ for the estimation of $P_\theta(T)$.
Specifically, $\mathrm{Train}$ is used to obtain the intermediate model parameters $\theta_t$ during training. 
$\mathrm{Push}$ is used to initialize $Q$ (\ie, $\mathrm{Push}$ $N$ times) and to maintain the diversity of models in $Q$ by periodically adding newly initialized models into $Q$. 
On one hand, this ensures the diversity of randomly initialized models whose parameters are sampled from $P_{\theta_0}$; On the other hand, this ensures the diversity of training iterations, \ie, the models in $Q$ always have different training iterations. 
$\mathrm{Pop}$ is used to control the size of $Q$ for the sake of computational cost. 

\vspace{2mm}
As shown in Alg.~\ref{algo}, after initialization, we loop through $\mathrm{Sample}$ and DM, $\mathrm{Train}$, $\mathrm{Push}$, $\mathrm{Pop}$, which implements the proposed memory-efficient model sampling scheme for data condensation.
Intuitively, we can visualize $P_{\theta}(T)$ as a matrix (Fig.~\ref{fig:framework}) with its rows as samples from $P_{\theta_0}$ and its columns as training iterations, and our scheme allows for a novel way of sampling $P_{\theta}(T)$ by {\bf ``scanning''} through the matrix. 
Unlike DM that only samples from the first column of the matrix and MTT~\cite{cazenavette2022dataset} that stores the entire matrix for sampling, our model queue is essentially a dynamic ``band'' that traverses the matrix from left to right and top to bottom, thereby achieving a better balance between $T$ and computational resources.
Specifically, the left-to-right traverse is implemented by $\mathrm{Train}$, and the top-to-bottom traverse is implemented by $\mathrm{Push}$ together with $\mathrm{Pop}$.

\vspace{2mm}
\noindent \textbf{Remark.}
Note that we implicitly assume that on average a model can only be trained for at most $K \times N_{\mathrm{max}}$ times before being popped from $Q$ and replaced by a newly initialized model.
This implies that $K \times N_{\mathrm{max}}$ should be large enough so that the trained models can extract all informative features of the real data for data condensation. Please see Sec.~\ref{sec:matching_with_trained} for an empirical verification of this claim.

\subsection{Class-aware Distribution Regularization}
\label{sec:class_aware_distribution_regularization}

As mentioned above, we enrich the model sampling in DM with an additional dimension ``training iterations'' and propose an efficient data structure, namely the memory-efficient model queue $Q$, to organize models with different training iterations at low computational costs.
In short, we have enriched $P_{\theta_0}$ in Eq.~\ref{eq:dm_objective} to $P_\theta(T)$. 

However, we observed that {\bf SC2} still holds, \ie, the enriched embeddings provided by $P_\theta(T)$ is still not enough for the MMD estimation.
Specifically, we observed that the extracted features are scattered around and only their means (\ie, first-order moment) are matched, which indicates that the minimization of MMD loss failed to match higher order moments of the real and synthetic distributions\cite{li2015generative} (Sec.~\ref{sec:class_misalignment}).
As a result, such scattered features are mixed with each other and become less distinguishable in terms of classification, which impairs the performance of data condensation.

Addressing this issue, we propose to add a classification loss (\ie, cross-entropy loss) as a regularization term to the synthetic distribution to make the extracted features more distinguishable. We conjecture this implicitly helps the matching of higher order moments of the two distributions\cite{li2015generative}.
However, since all our models are sampled from the model queue $Q$ and thus have different training iterations, their classification accuracy varies and we should only require the synthetic data to achieve the same classification accuracy as the real data.
To this end, let $\phi$ be a model sampled from $Q$, $Acc_{\phi}$ be its accuracy on real data, $\mathcal{L}_{CE}$ be the cross-entropy loss, we have:
\begin{equation}
    \arg \min_{\mathcal{S}} Acc_{\phi} \mathcal{L}_{CE}(\mathcal{S}),
    \label{eq:ce_reg}
\end{equation}
\noindent \textbf{Remark.}
We did not attempt to improve MMD as it is too challenging to design a neural network based mapping function with tractable kernels.
We also abandoned the clustering approach (\ie, clustering features of the same class towards their mean) as it is less relevant to the classification task and thus less effective for dataset condensation.

\subsection{Overall Loss Function and Pseudocode}
\label{sec:overall_loss}

In summary, the overall loss function of our method is as follows\footnote{Standard cross-entropy loss is used to train models in the model queue.}:
\begin{equation}
    \mathcal{L}_{overall} = \mathcal{L}_{DM_{\phi \sim P_{\theta}(T)}} + \lambda_{reg} Acc_{\phi} \mathcal{L}_{CE},
    \label{eq:overall_loss}
\end{equation}
where $\mathcal{L}_{DM_{\phi \sim P_{\theta}(T)}}$ is the modified version of the distribution matching loss depicted in Eq.~\ref{eq:dm_objective} which samples model parameter $\phi$ from our enriched distribution $P_{\theta}(T)$, $Acc_{\phi} \mathcal{L}_{CE}$ is our class-aware distribution regularizer with a weighting parameter $\lambda_{reg}$. The pseudocode of our algorithm is shown in Alg.~\ref{algo}.

\begin{algorithm}
\SetKwInOut{Params}{Params}
\KwIn{Training set $\mt$}

\Params{$\phi$: network; $P_{\btheta_0}$: distribution of randomly initialized network weights; $\bm{Q}$: model queue; $N$: initial queue size; $N_{max}$: maximum queue size; $K$: step number; $M$: training iterations.}

Initialize $Q$ with $N$ randomly initialized models $\phi_{\btheta_i}$ that $\btheta_i \sim P_{\btheta_0}$.

\For{$m = 0, \cdots, M-1$}
{
        Sample $\phi_{\btheta}$ and from $\bm{Q}$ and calculate its $Acc_{\phi_{\btheta}}$
        
        \tcc{Improved DM}
        
        Calculate $\mathcal{L}_{DM}$ using Eq.~\ref{eq:dm_objective} but with $\phi_{\btheta}$

        Calculate $\mathcal{L}_{CE}$ according to Eq.~\ref{eq:ce_reg}

        Update condensed set $\ms$ by minimizing Eq.~\ref{eq:overall_loss}

        \tcc{Model Queue $\bm{Q}$ Maintenance}
        
        Train $\phi_{\btheta}$ for $K$ steps to $\phi_{\btheta'}$ and put back in $\bm{Q}$

        Push a new model $\phi_{\btheta_0}$ that $\btheta_0 \sim P_{\btheta_0}$ to $\bm{Q}$

        Pop a model from $\bm{Q}$ if $|\bm{Q}| > N_{max}$
}
\KwOut{Condensed set $\ms$}
\caption{Improved Distribution Matching for Dataset Condensation}
\label{algo}
\end{algorithm}

\section{Experiment}

\subsection{Experimental Setup}

Following the evaluation protocol of previous dataset condensation studies, we use image classification as a proxy task for evaluation and report the classification accuracy of deep neural networks trained on the condensed set synthesized by our method. 

\vspace{2mm}
\noindent \textbf{Datasets.} 
Given the saturating performance of dataset condensation (DC) on simple datasets like MNIST\cite{lecun1998gradient} and FashionMNIST\cite{xiao2017fashion}, we evaluate our method on four larger and more complex datasets, including CIFAR-10, CIFAR-100\cite{krizhevsky2009learning}, TinyImageNet\cite{le2015tiny} and a subset of ImageNet\cite{deng2009imagenet}. 
Following \cite{tian2020contrastive}, ImageNet Subset selects 100 categories from the ImageNet dataset, whose high-resolution images ($224\times224$) contain more realistic patterns and are thus closer to real-world application scenarios. 

\vspace{2mm}
\noindent \textbf{Experiment Settings.} Following previous studies\cite{zhao2021dataset,zhao2021DM,zhao2021DSA}, we evaluate DC methods with three different settings for each dataset: condensing it to different synthetic sets of 1, 10, and 50 images per class, respectively. 
Unless specified, we follow DM\cite{zhao2021DM} and use the same ConvNet architecture\cite{gidaris2018dynamic} in all the experiments on CIFAR-10, CIFAR-100 and TinyImageNet. 
This ConvNet consists of three blocks and each block is made up of a 128-kernel convolution layer, an instance normalization layer\cite{ulyanov2016instance}, a ReLU activation function\cite{nair2010rectified}, and an average pooling layer.
The instance normalization is used to facilitate training on small condensed sets.
For ImageNet Subset, we increase the number of blocks in the abovementioned ConvNet to 6 to cope with its higher resolution and more complex patterns.
For the evaluation, we use the same network architectures used in DC and report the mean accuracy and standard deviation of 5 runs where the models are randomly initialized and trained for 1000 epochs using the condensed set. 

\vspace{2mm}
\noindent \textbf{Implementation Details.} 
We follow the implementation of DM\cite{zhao2021DM} to set most hyper-parameters of our method.
Specifically, for model training, we use the same SGD optimizer setting in both DC and evaluation, where the learning rate is 0.01, momentum is 0.9 and weight decay is 0.0005; for the optimization of the condensed set, we use a learning rate of 0.2 with a momentum of 0.5.
For the different experimental settings mentioned above, we set different $\lambda_{reg}$ for the cross-entropy loss, \ie, 0.5 for the cases of 1 and 10 condensed images per class, and 0.1 for the case of 50 images per class. 
The hyper-parameters of Model Parameter Sampling are set according to the computation demand for different datasets. For CIFAR-10/100, we set the maximum queue size $N_{max}$ as 100 and the training step $K$ as 10.
Following DM, we perform distribution matching for each class separately.
We initialize the condensed set by randomly sampling images from the training set. The partition number $l$ is set as 2.
We run all our experiments in Table~\ref{table:main_table} with a single GTX 3090 GPU with 24GB memory.
Please see the supplementary material for more details.

\begin{table*}
\centering
\caption{Comparison with previous coreset selection and dataset condensation methods. As in previous work, we evaluate our method on four different datasets with different numbers of synthetic images per class. Img/Cls: number of images per class. Ratio(\%): the ratio of condensed images to the whole training set. Whole Dataset: the accuracy of the model trained on the whole training set. 
Note: DD\textsuperscript{\textdagger} and LD\textsuperscript{\textdagger} use different architectures \ie LeNet\cite{lecun1998gradient} for MNIST and AlexNet\cite{krizhevsky2017imagenet} for CIFAR10. The rest of the methods all use ConvNet\cite{gidaris2018dynamic}.}
\label{table:main_table}
\resizebox{2.\columnwidth}{!}{
\begin{tabular}{c|ccc|ccc|ccc|ccc} 
\toprule
              & \multicolumn{3}{c|}{CIFAR-10}  & \multicolumn{3}{c|}{CIFAR100}   & \multicolumn{3}{c|}{TinyImageNet} & \multicolumn{3}{c}{ImageNet Subset}      \\ 
\midrule
Img/Cls       & 1        & 10       & 50       & 1        & 10       & 50        & 1        & 10       & 50          & 1        & 10       & 50                 \\
Ratio (\%)    & 0.02     & 0.2      & 1        & 0.2      & 2        & 10        & 0.2      & 2        & 10          & 0.08      & 0.8        & 4                 \\ 
\midrule
Random        & 14.4±2.0 & 26.0±1.2 & 43.4±1.0 & 4.2±0.3  & 14.6±0.5 & 30.0±0.4  & 1.4±0.1  & 5.0±0.2  & 15.0±0.4    & 2.4±0.3  & 6.2±0.0  & 10.0±0.1           \\
Herding       & 21.5±1.2 & 31.6±0.7 & 40.4±0.6 & 8.4±0.3  & 17.3±0.3 & 33.7±0.5  & 2.8±0.2  & 6.3±0.2  & 16.7±0.3    & 3.0±0.2  & 8.3±0.1 & 14.8±0.4           \\
Forgetting    & 13.5±1.2 & 23.3±1.0 & 23.3±1.1 & 4.5±0.2  & 15.1±0.3 & 30.5±0.3  & 1.6±0.1  & 5.1±0.2  & 15.0±0.3    & 1.4±0.2  & 4.5±0.4 & 9.0±0.6           \\ 
\midrule
DD            & -        & 36.8±1.2 & -        & -        & -        & -         & -        & -        & -           & -        & -        & -                  \\
LD            & 25.7±0.7 & 38.3±0.4 & 42.5±0.4 & 11.5±0.4 & -        & -         & -        & -        & -           & -        & -        & -                  \\ 
\midrule
DC            & 28.3±0.5 & 44.9±0.5 & 53.9±0.5 & 12.8±0.3 & 25.2±0.3 & -         & -        & -        & -           & -        & -        & -                  \\
DSA           & 28.8±0.7 & 52.1±0.5 & 60.6±0.5 & 13.9±0.3 & 32.3±0.3 & 42.8±0.4  & -        & -        & -           & -        & -        & -                  \\
CAFE          & 30.3±1.1 & 46.3±1.6 & 55.5±0.6 & 12.9±0.3 & 27.8±0.3 & 37.9±0.3  & -        & -        & -           & -        & -        & -                  \\
CAFE+DSA      & 31.6±0.8 & 50.9±0.5 & 62.3±0.4 & 14.0±0.3 & 31.5±0.2 & 42.9±0.2  & -        & -        & -           & -        & -        & -                  \\
DM            & 26.0±0.8 & 48.9±0.6 & 63.0±0.4 & 11.4±0.3 & 29.7±0.3 & 43.6±0.4  & 3.9±0.2  & 12.9±0.4 & 24.1±0.3    & 4.5±0.4        & 11.9±0.3 & 22.5±0.3           \\ 
\midrule
IDM (Ours)  & \textbf{45.6±0.7} & \textbf{58.6±0.1} & \textbf{67.5±0.1} & \textbf{20.1±0.3} & \textbf{45.1±0.1} & \textbf{50.0±0.2}  & \textbf{10.1±0.2} & \textbf{21.9±0.2} & \textbf{27.7±0.3}                 & \textbf{11.2±0.5} & \textbf{17.1±0.6} & \textbf{26.3±0.4}           \\ 
\midrule
Whole Dataset & \multicolumn{3}{c|}{84.8±0.1}                    & \multicolumn{3}{c|}{56.2±0.3}     & \multicolumn{3}{c|}{37.6±0.4}  & \multicolumn{3}{c}{46.8±0.6}             \\
\midrule
\end{tabular}
}
\end{table*}

\subsection{Comparison with Previous Methods}

Table~\ref{table:main_table} shows the comparison of our method with previous coreset selection and data condensation methods.
Following previous studies, we compare our method to three {\it coreset selection} methods, Random, Herding\cite{rebuffi2017icarl,castro2018end}, and Forgetting\cite{toneva2018empirical}. Specifically, Random refers to randomly selects images from the training set $\mathcal{T}$ as the condensed set $\mathcal{S}$; Herding selects the samples in $\mathcal{T}$ that are closest to the clustering center for each class as $\mathcal{S}$; Forgetting selects the samples in $\mathcal{T}$ that are more frequently forgotten during the model training as $\mathcal{S}$. 
For previous {\it dataset condensation} methods, we compare to: 
DD\cite{wang2018dataset} and LD\cite{bohdal2020flexible}, two pioneering dataset condensation methods whose performance are evaluated on different architectures;
DC\cite{zhao2021dataset}, DSA\cite{zhao2021DSA} and CAFE\cite{wang2022cafe}, three typical optimization-oriented dataset condensation methods that achieved significant performance improvement on small and lower-resolution datasets;
DM\cite{zhao2021DM}, the first method that approached data condensation via distribution matching, which works as our baseline. 
Following DM\cite{zhao2021DM}, we do not provide the results of optimization-oriented method on TinyImageNet and ImageNet Subset due to limited computational resources.

The improvement of our IDM over DM is significant.
For example, our IDM surpasses DM by 9.7\% in CIFAR-10 10 Img/Cls and 15.4\% in CIFAR-100 10 Img/Cls.
As a result, unlike DM that lags behind most optimization-oriented methods (\eg, DC, DSA, CAFE), our IDM outperforms all of them, showing the competitiveness of distribution matching based methods and thus shedding light on follow-up research.
In addition, thanks to our partitioning and expansion augmentation, our IDM produces higher numbers of features, leading to better distribution estimations and greater improvements over DM in challenging settings (\eg, CIFAR-10 1 Img/Cls).
For example, we achieve a 19.6\% performance improvement over DM in CIFAR-10 1 Img/Cls.
For TinyImageNet and ImageNet Subset, IDM achieved similar performance improvements.

\subsection{Justification of Algorithmic Motivation}

\noindent \textbf{Trained Models vs. Random Models.} 
\label{sec:matching_with_trained}
We would like to point out that an important conclusion of DM\cite{zhao2021DM}, which claims that the use of trained models does not significantly improve the performance of dataset condensation, only applies to small and simple datasets like CIFAR-10 and does not generalize to larger datasets like CIFAR-100.

\begin{figure}[t]
  \centering
   \includegraphics[width=0.65\linewidth]{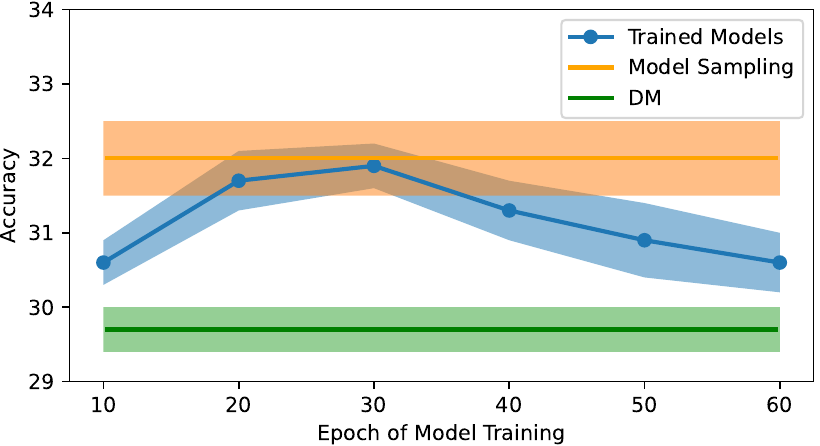}

   \caption{The performance of DM using models trained for different numbers of epochs on CIFAR-100.}
   \label{fig:weakly_trained}
\end{figure}

To support our claim, we first pre-train 600 ConvNets using the entire training set of CIFAR-100 for 10, 20, 30, 40, 50, and 60 epochs respectively.
Then we perform DM using models trained with the same number of epochs for 10 Img/Cls condensation. 
As Fig.~\ref{fig:weakly_trained} shows, i) the performance of DM using pre-trained models significantly outperform that using random models (\ie, DM) across all six settings; ii) our model sampling strategy, using both trained and random models, achieves the best performance. 
Note that our model sampling avoids the space-consuming preparation of trained models and the time-consuming tuning of the number of training epochs.

\begin{figure}[t]
  \centering
   \includegraphics[width=0.6\linewidth]{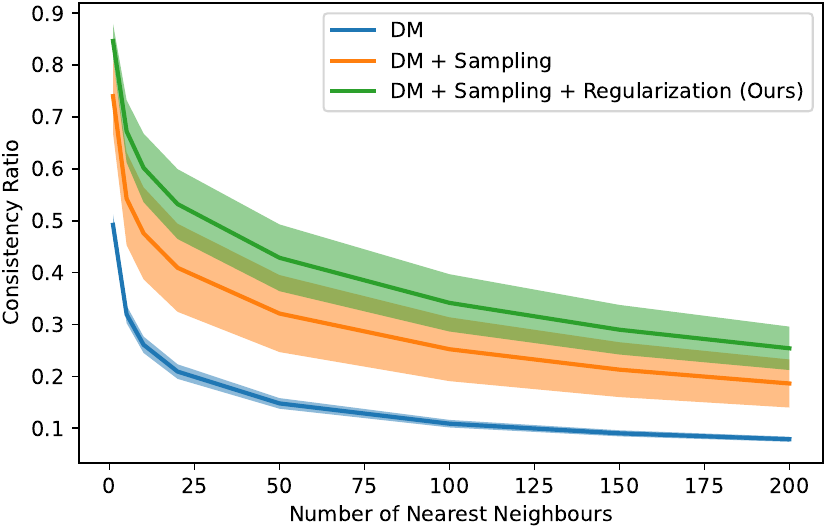}

    \caption{The proposed model sampling (Sec.~\ref{sec:efficient_enriched_sampling}) and regularization (Sec.~\ref{sec:class_aware_distribution_regularization}) techniques effectively alleviate DM's class misalignment problem. 
    $y$-axis: consistency ratio, \ie, the ratio of each synthetic sample's real image neighbours that belong to the same class of the synthetic sample in the feature space. $x$-axis: the number of nearest neighbours used in consistency ratio calculation.}
   \label{fig:misalignment}
   \vspace{-4mm}
\end{figure}

\vspace{2mm}
\noindent \textbf{Class Misalignment.}
\label{sec:class_misalignment}
To justify that our IDM effectively alleviate the class misalignment problem of DM\cite{zhao2021DM}, we report the ``consistency ratio'', which is the ratio of each synthetic sample's real image neighbours that belong to the same class of the synthetic sample in the feature space, on three settings: DM, DM + our model sampling (Sec.~\ref{sec:efficient_enriched_sampling}), and DM + our model sampling + our regularization (Sec.~\ref{sec:class_aware_distribution_regularization}), \ie, Ours. 
Specifically, we use $k$-nearest neighbors algorithm ($k$-NN) to get $k$ real image neighbours and L2 norm as the similarity metric following the distribution matching loss.
Fig.~\ref{fig:misalignment} shows the experimental results on CIFAR-100 with 10 Img/Cls. 
It can be observed that both our model sampling and regularization improve the consistency ratio with different $k$, suggesting that our IDM has effectively alleviated DM's class misalignment problem.

\subsection{Ablation Study}

\vspace{2mm}
\noindent \textbf{Effectiveness of Each Component.}
As Table~\ref{table:ablation} shows, the three components of our method, \ie, model sampling (Sec.~\ref{sec:efficient_enriched_sampling}), distribution regularization (Sec.~\ref{sec:class_aware_distribution_regularization}) and augmentation (Sec.~\ref{sec:augmentation}) improve the performance of DM on CIFAR-100 by 2.3\%, 2.3\% and 5.4\% respectively. Interestingly, our model sampling lowers the performance of DM on ImageNet Subset by 1.1\% while our distribution regularization improves the performance by a large margin of 3.7\%. 
We ascribe this to our limited computational resources against large datasets, which greatly reduces the number of models affordable in our model queue, making them insufficient for training.
Please note that distribution regularization is infeasible without model sampling.

\begin{figure*}[!htb]
\minipage{0.28\textwidth}
  \includegraphics[width=\linewidth]{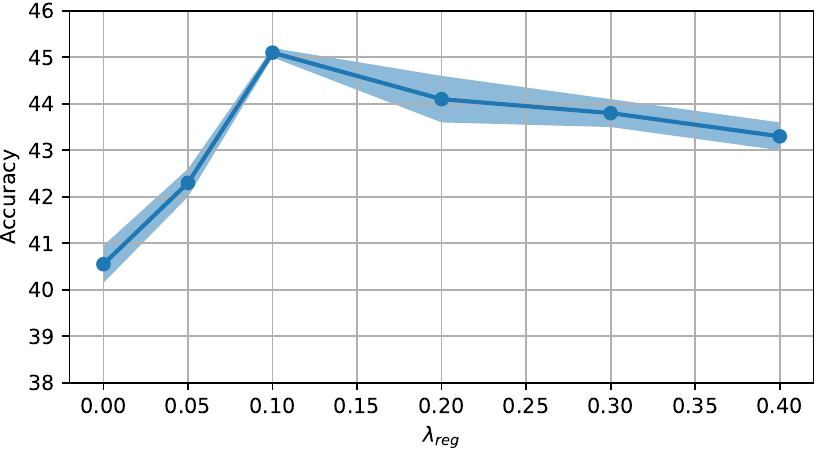}
  \caption{Ablation of $\lambda_{reg}$ (Sec.~\ref{sec:overall_loss}).}
  \label{fig:sensitivity_ceweight}
\endminipage\hfill
\minipage{0.28\textwidth}
  \includegraphics[width=\linewidth]{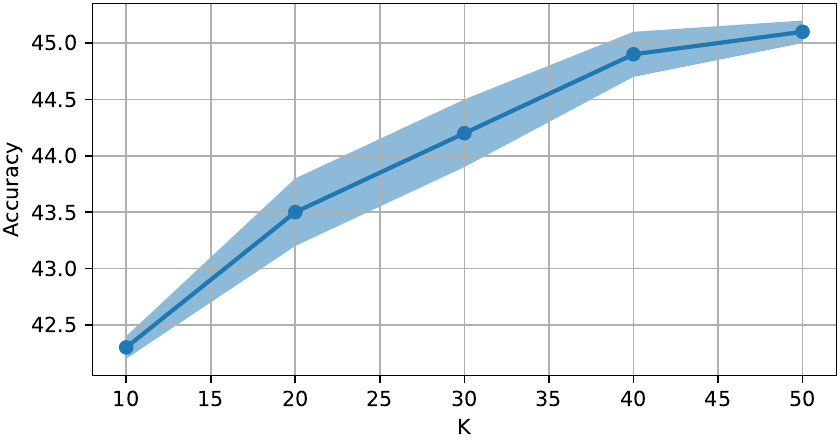}
  \caption{Ablation of $K$ (Sec.~\ref{sec:efficient_enriched_sampling}).}
  \label{fig:sensitivity_trainstep}
\endminipage\hfill
\minipage{0.28\textwidth}%
  \includegraphics[width=\linewidth]{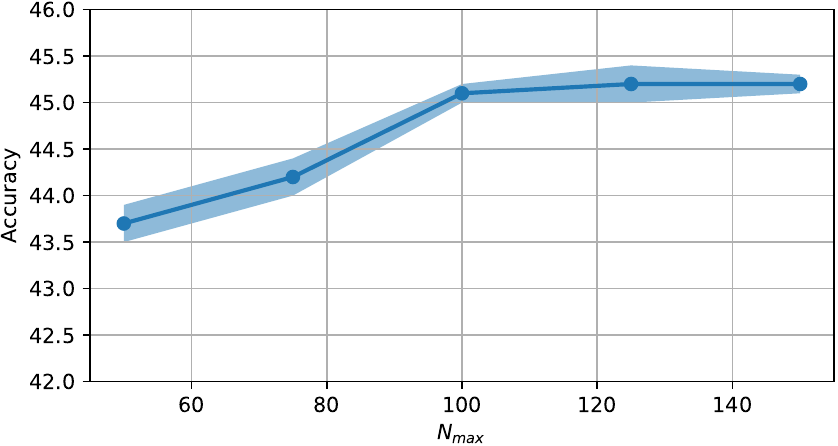}
  \caption{Ablation of $N_{\mathrm{max}}$ (Sec.~\ref{sec:efficient_enriched_sampling}).}
  \label{fig:sensitivity_queuesize}
\endminipage
\vspace{-4mm}
\end{figure*}

\vspace{2mm}
\noindent \textbf{Sensitivity of Hyper-parameters.}
Fig.~\ref{fig:sensitivity_ceweight}, Fig.~\ref{fig:sensitivity_trainstep} and Fig.~\ref{fig:sensitivity_queuesize} show how the performance of our IDM change with different choices of regularization weight $\lambda_{reg}$ (Sec.~\ref{sec:overall_loss}), the number of steps a model is trained in each training iteration $K$ (Sec.~\ref{sec:efficient_enriched_sampling}), and the size of the model queue $N_{\mathrm{max}}$ (Sec.~\ref{sec:efficient_enriched_sampling}), respectively. 
The experiments are conducted on CIFAR-100 with 10 Img/Cls.
Note that the best performance achieved are higher than those in Table~\ref{table:main_table} because we further optimized the choices of hyper-parameters.

\begin{table}
\centering
\caption{Ablation study on CIFAR-100 and ImageNet Subset.}
\label{table:ablation}
\resizebox{0.85\columnwidth}{!}{
\begin{tabular}{l|c|c} 
\toprule
Dataset                 & CIFAR100 & ImageNet Subset \\ 
\midrule
Img/Cls & 10 & 10 \\
\midrule
DM      				& 29.7±0.3        & 11.9±0.3              \\
+ Model Sampling    & 32.0±0.5     & 10.8±0.7             \\ 
+ Distribution Regularization        & 34.3±0.3 & 15.6±0.4        \\
+ Augmentation        & 45.1±0.1 & 17.1±0.6        \\
\midrule
\end{tabular}
}
\end{table}

\vspace{2mm}
\noindent \textbf{Number of Partition $l^2$ in Augmentation (Sec.~\ref{sec:augmentation}).} 
As Table~\ref{table:partition} shows, using a partition of $2 \times 2$ achieves the best performance on ImageNet Subset with 1 Img/Cls. The performance of $3 \times 3$ partition is slightly worse as it discarded too much image details in each partition.
This indicates that partitions with high $l$ are not applicable to datasets of low-resolution images (\eg, CIFAR-10/100). 

\begin{table}
\centering
\caption{Ablation study of the number of partition $l^2$ in our augmentation (Sec.~\ref{sec:augmentation}) on ImageNet Subset with 1 Img/Cls.}
\label{table:partition}
\resizebox{0.65\columnwidth}{!}{
\begin{tabular}{l|c|c|c} 
\toprule
Partition ($l \times l$) & 1$\times$1 & 2$\times$2 & 3$\times$3 \\ 
\midrule
Accuracy & 5.4±0.1 & 11.2±0.5 & 10.7±0.7 \\
\midrule
\end{tabular}
}
\vspace{-4mm}
\end{table}

\subsection{Ablation Study on CIFAR-10 Architectural Generalization}

Following previous studies\cite{zhao2021dataset,zhao2021DM}, we verify the cross-architectural transferability of the condensed sets on CIFAR-10 with 10 Img/Cls. Specifically, we perform data condensation with one architecture (denoted as C) and evaluate  the effectiveness of the condensed set obtained with another architecture (denoted as T). 
Following DM\cite{zhao2021DM}, we evaluate our method on four different architectures, including ConvNet, AlexNet\cite{krizhevsky2017imagenet}, VGG11\cite{vgg} and ResNet18\cite{he2016deep}. The experiments setting and all hyper-parameters are the same as Table~\ref{table:main_table}, except that we set the partition parameter $l=1$ to allow for a fair comparison with DM.
As Table~\ref{table:cross_arch} shows, i) dataset condensation using our IDM method and ConvNet performs the best on all four evaluation architectures and significantly outperforms those of DM; ii) similar to the discussion in DM, dataset condensation with complex architectures generally performs worse due to the difficulties in optimization and noisy extracted features.
A detailed comparison is provided in the supplementary material.

\begin{table}
\centering
\caption{Cross-architectural performance of our IDM method on CIFAR-10 with 10 Img/Cls. Our IDM achieves a significant improvement over DM.}
\label{table:cross_arch}
\resizebox{0.85\columnwidth}{!}{
\begin{tabular}{l|c|c|c|c|c} 
\toprule
                        & C$\backslash$T    & ConvNet  & AlexNet & VGG & ResNet  \\ 
\midrule
DM                     & ConvNet & 48.9±0.6       & 38.8±0.5  & 42.1±0.4 & 41.2±1.1  \\
\midrule
\multirow{4}{*}{Ours}     & ConvNet & 53.0±0.3 & 44.6±0.8 & 47.8±1.1 & 44.6±0.4        \\
                        & AlexNet & 44.8±0.5 & 41.4±1.4        & 43.1±0.6    & 41.0±0.1        \\
                        & VGG     & 41.2±0.4 & 37.4±0.3        & 41.7±0.4    &  38.8±0.8       \\ 
                        & ResNet  & 38.3±0.4 & 37.0±0.7        & 39.0±0.1    & 39.0±0.4        \\
\midrule
\end{tabular}
}
\vspace{-3mm}
\end{table}

\begin{figure}[t!]
    \centering
    \begin{subfigure}[t]{0.49\linewidth}
        \centering
        \includegraphics[width=\linewidth]{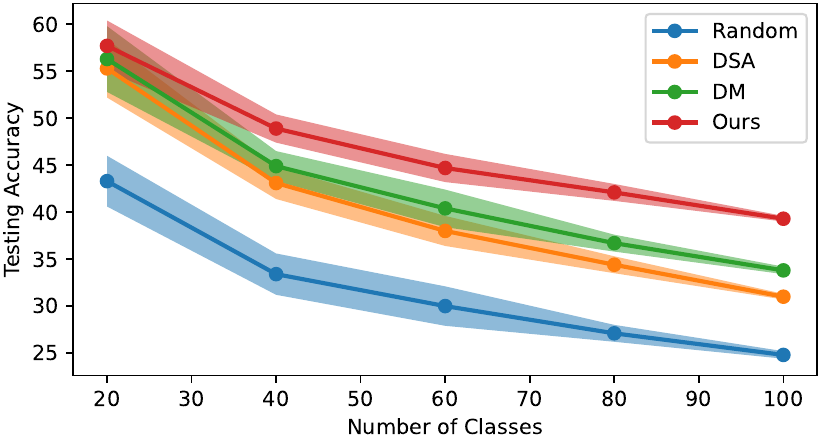}
        \caption{}
        \label{fig:continual_5step}
    \end{subfigure}%
    ~ 
    \begin{subfigure}[t]{0.49\linewidth}
        \centering
        \includegraphics[width=\linewidth]{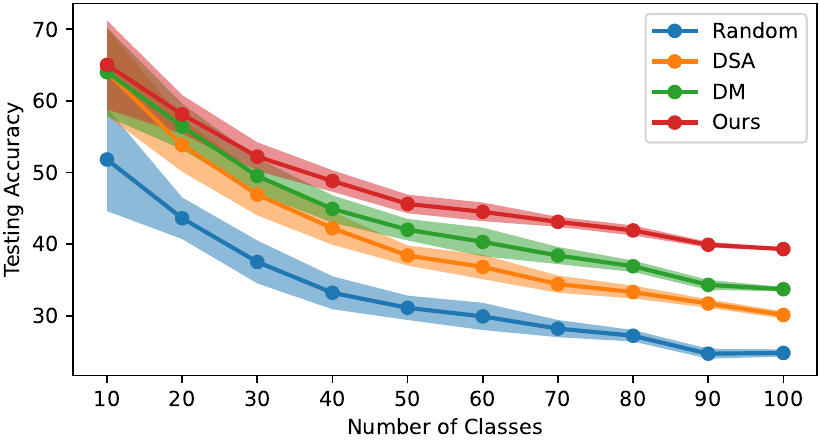}
        \caption{}
        \label{fig:continual_10step}
    \end{subfigure}
    \caption{Application in continual learning. (a) 5-step and (b) 10-step continual learning on CIFAR-100.}
    \vspace{-4mm}
\end{figure}

\subsection{Continual Learning}

Dataset condensation facilitates continual learning\cite{rebuffi2017icarl} by alleviating its catastrophic forgetting problem by storing more efficient training samples in the memory.
Following previous studies, we build a baseline based on GDumb\cite{prabhu2020gdumb} which stores training samples greedily in the memory and maintains their class-balance. 
The model is trained from scratch on the latest memory only to make the continual learning performance a valid metric to measure the quality of memory construction and condensed data. 
Following DM\cite{zhao2021DM}, we compare our method with three competitors, including Random, DSA\cite{zhao2021DSA} and DM. The continual learning experiments are conducted on CIFAR-100 with 5-step and 10-step settings, where the ``step'' indicates the number of stages in the continual training. All the experiments are repeated five times with different class order in continual learning, and we then report the mean and standard deviation of performance. The synthetic image budget for the experiment is 20 Img/Cls following DM, and we use the same class order to DM in every step to make the result comparable. The condensed images are synthesized with the same hyper-parameters of the experiment of CIFAR-100 with 10 Img/Cls, except that we set the partition parameter $l=1$ to allow for a fair comparison with other methods.

As shown in Fig.~\ref{fig:continual_5step} and Fig.~\ref{fig:continual_10step}, our method outperforms Random, DSA and DM in both settings, indicating that the condensed data generated by our method are of the highest quality for continual learning. 
The final performance of our method is 39.3\%/39.3\% for 5/10 steps, while the performance of Random, DSA and DM are 24.8\%/25.2\%, 31.0\%/29.8\% and 33.8\%/33.7\%, respectively.

\section{Conclusion}

We propose an improved distribution matching (IDM) method for dataset condensation by alleviating two important shortcomings of naive distribution matching, \ie, imbalanced feature numbers and unvalidated embeddings for distance computation, with three novel techniques, \ie, partitioning and expansion augmentation, efficient and enriched model sampling, and class-aware distribution regularization.
As a result, our method achieves significant improvements over previous methods while requiring fewer computational resources. 
This also allows our method to be applied to larger datasets with more categories with minimal extra cost. 
Extensive experiments demonstrate the effectiveness of our method, showing the competitiveness of distribution matching based methods and thus shedding light on follow-up research.

\noindent \textbf{Acknowledgments}
This work was supported 
by the Guangdong Basic and Applied Basic Research Foundation (NO. 2020B1515020048), 
the National Natural Science Foundation of China (NO. 61976250), 
the Shenzhen Science and Technology Program (NO. JCYJ20220530141211024) and 
the Fundamental Research Funds for the Central Universities under Grant 22lgqb25.

{\small
\bibliographystyle{ieee_fullname}
\bibliography{egbib}

\begin{thebibliography}{10}\itemsep=-1pt

\bibitem{agarwal2004approximating}
Pankaj~K Agarwal, Sariel Har-Peled, and Kasturi~R Varadarajan.
\newblock Approximating extent measures of points.
\newblock {\em Journal of the ACM (JACM)}, 51(4):606--635, 2004.

\bibitem{aljundi2019gradient}
Rahaf Aljundi, Min Lin, Baptiste Goujaud, and Yoshua Bengio.
\newblock Gradient based sample selection for online continual learning.
\newblock {\em Advances in neural information processing systems}, 32, 2019.

\bibitem{amid2022learning}
Ehsan Amid, Rohan Anil, Wojciech Kot{\l}owski, and Manfred~K Warmuth.
\newblock Learning from randomly initialized neural network features.
\newblock {\em arXiv preprint arXiv:2202.06438}, 2022.

\bibitem{bohdal2020flexible}
Ondrej Bohdal, Yongxin Yang, and Timothy Hospedales.
\newblock Flexible dataset distillation: Learn labels instead of images.
\newblock {\em arXiv preprint arXiv:2006.08572}, 2020.

\bibitem{cao2018review}
Weipeng Cao, Xizhao Wang, Zhong Ming, and Jinzhu Gao.
\newblock A review on neural networks with random weights.
\newblock {\em Neurocomputing}, 275:278--287, 2018.

\bibitem{caron2018deep}
Mathilde Caron, Piotr Bojanowski, Armand Joulin, and Matthijs Douze.
\newblock Deep clustering for unsupervised learning of visual features.
\newblock In {\em Proceedings of the European conference on computer vision
  (ECCV)}, pages 132--149, 2018.

\bibitem{castro2018end}
Francisco~M Castro, Manuel~J Mar{\'\i}n-Jim{\'e}nez, Nicol{\'a}s Guil, Cordelia
  Schmid, and Karteek Alahari.
\newblock End-to-end incremental learning.
\newblock In {\em Proceedings of the European conference on computer vision
  (ECCV)}, pages 233--248, 2018.

\bibitem{cazenavette2022dataset}
George Cazenavette, Tongzhou Wang, Antonio Torralba, Alexei~A Efros, and
  Jun-Yan Zhu.
\newblock Dataset distillation by matching training trajectories.
\newblock In {\em Proceedings of the IEEE/CVF Conference on Computer Vision and
  Pattern Recognition}, pages 4750--4759, 2022.

\bibitem{chen2012super}
Yutian Chen, Max Welling, and Alex Smola.
\newblock Super-samples from kernel herding.
\newblock {\em arXiv preprint arXiv:1203.3472}, 2012.

\bibitem{deng2009imagenet}
Jia Deng, Wei Dong, Richard Socher, Li-Jia Li, Kai Li, and Li Fei-Fei.
\newblock Imagenet: A large-scale hierarchical image database.
\newblock In {\em 2009 IEEE conference on computer vision and pattern
  recognition}, pages 248--255. Ieee, 2009.

\bibitem{dong2022privacy}
Tian Dong, Bo Zhao, and Lingjuan Lyu.
\newblock Privacy for free: How does dataset condensation help privacy?
\newblock {\em arXiv preprint arXiv:2206.00240}, 2022.

\bibitem{feldman2020introduction}
Dan Feldman.
\newblock Introduction to core-sets: an updated survey.
\newblock {\em arXiv preprint arXiv:2011.09384}, 2020.

\bibitem{feldman2011scalable}
Dan Feldman, Matthew Faulkner, and Andreas Krause.
\newblock Scalable training of mixture models via coresets.
\newblock {\em Advances in neural information processing systems}, 24, 2011.

\bibitem{feldman2007ptas}
Dan Feldman, Morteza Monemizadeh, and Christian Sohler.
\newblock A ptas for k-means clustering based on weak coresets.
\newblock In {\em Proceedings of the twenty-third annual symposium on
  Computational geometry}, pages 11--18, 2007.

\bibitem{feldman2020turning}
Dan Feldman, Melanie Schmidt, and Christian Sohler.
\newblock Turning big data into tiny data: Constant-size coresets for k-means,
  pca, and projective clustering.
\newblock {\em SIAM Journal on Computing}, 49(3):601--657, 2020.

\bibitem{gidaris2018dynamic}
Spyros Gidaris and Nikos Komodakis.
\newblock Dynamic few-shot visual learning without forgetting.
\newblock In {\em Proceedings of the IEEE conference on computer vision and
  pattern recognition}, pages 4367--4375, 2018.

\bibitem{goetz2020federated}
Jack Goetz and Ambuj Tewari.
\newblock Federated learning via synthetic data.
\newblock {\em arXiv preprint arXiv:2008.04489}, 2020.

\bibitem{gretton2012kernel}
Arthur Gretton, Karsten~M Borgwardt, Malte~J Rasch, Bernhard Sch{\"o}lkopf, and
  Alexander Smola.
\newblock A kernel two-sample test.
\newblock {\em The Journal of Machine Learning Research}, 13(1):723--773, 2012.

\bibitem{har2004coresets}
Sariel Har-Peled and Soham Mazumdar.
\newblock On coresets for k-means and k-median clustering.
\newblock In {\em Proceedings of the thirty-sixth annual ACM symposium on
  Theory of computing}, pages 291--300, 2004.

\bibitem{he2016deep}
Kaiming He, Xiangyu Zhang, Shaoqing Ren, and Jian Sun.
\newblock Deep residual learning for image recognition.
\newblock In {\em Proceedings of the IEEE conference on computer vision and
  pattern recognition}, pages 770--778, 2016.

\bibitem{pmlr-v162-kim22c}
Jang-Hyun Kim, Jinuk Kim, Seong~Joon Oh, Sangdoo Yun, Hwanjun Song, Joonhyun
  Jeong, Jung-Woo Ha, and Hyun~Oh Song.
\newblock Dataset condensation via efficient synthetic-data parameterization.
\newblock In {\em Proceedings of the 39th International Conference on Machine
  Learning}, pages 11102--11118, 2022.

\bibitem{krizhevsky2009learning}
Alex Krizhevsky, Geoffrey Hinton, et~al.
\newblock Learning multiple layers of features from tiny images.
\newblock 2009.

\bibitem{krizhevsky2017imagenet}
Alex Krizhevsky, Ilya Sutskever, and Geoffrey~E Hinton.
\newblock Imagenet classification with deep convolutional neural networks.
\newblock {\em Communications of the ACM}, 60(6):84--90, 2017.

\bibitem{le2015tiny}
Ya Le and Xuan Yang.
\newblock Tiny imagenet visual recognition challenge.
\newblock {\em CS 231N}, 7(7):3, 2015.

\bibitem{lecun1998gradient}
Yann LeCun, L{\'e}on Bottou, Yoshua Bengio, and Patrick Haffner.
\newblock Gradient-based learning applied to document recognition.
\newblock {\em Proceedings of the IEEE}, 86(11):2278--2324, 1998.

\bibitem{lee2022dataset}
Saehyung Lee, Sanghyuk Chun, Sangwon Jung, Sangdoo Yun, and Sungroh Yoon.
\newblock Dataset condensation with contrastive signals.
\newblock {\em Proceedings of the 39th International Conference on Machine
  Learning}, 2022.

\bibitem{li2020soft}
Guang Li, Ren Togo, Takahiro Ogawa, and Miki Haseyama.
\newblock Soft-label anonymous gastric x-ray image distillation.
\newblock In {\em 2020 IEEE International Conference on Image Processing
  (ICIP)}, pages 305--309. IEEE, 2020.

\bibitem{li2015generative}
Yujia Li, Kevin Swersky, and Rich Zemel.
\newblock Generative moment matching networks.
\newblock In {\em International conference on machine learning}, pages
  1718--1727. PMLR, 2015.

\bibitem{masarczyk2020reducing}
Wojciech Masarczyk and Ivona Tautkute.
\newblock Reducing catastrophic forgetting with learning on synthetic data.
\newblock In {\em Proceedings of the IEEE/CVF Conference on Computer Vision and
  Pattern Recognition Workshops}, pages 252--253, 2020.

\bibitem{nair2010rectified}
Vinod Nair and Geoffrey~E Hinton.
\newblock Rectified linear units improve restricted boltzmann machines.
\newblock In {\em Proceedings of the 27th International Conference on
  International Conference on Machine Learning}, pages 807--814, 2010.

\bibitem{nguyen2020dataset}
Timothy Nguyen, Zhourong Chen, and Jaehoon Lee.
\newblock Dataset meta-learning from kernel ridge-regression.
\newblock In {\em International Conference on Learning Representations}, 2020.

\bibitem{nguyen2021dataset}
Timothy Nguyen, Roman Novak, Lechao Xiao, and Jaehoon Lee.
\newblock Dataset distillation with infinitely wide convolutional networks.
\newblock {\em Advances in Neural Information Processing Systems},
  34:5186--5198, 2021.

\bibitem{nichol2018first}
Alex Nichol, Joshua Achiam, and John Schulman.
\newblock On first-order meta-learning algorithms.
\newblock {\em arXiv preprint arXiv:1803.02999}, 2018.

\bibitem{prabhu2020gdumb}
Ameya Prabhu, Philip~HS Torr, and Puneet~K Dokania.
\newblock Gdumb: A simple approach that questions our progress in continual
  learning.
\newblock In {\em European conference on computer vision}, pages 524--540.
  Springer, 2020.

\bibitem{rebuffi2017icarl}
Sylvestre-Alvise Rebuffi, Alexander Kolesnikov, Georg Sperl, and Christoph~H
  Lampert.
\newblock icarl: Incremental classifier and representation learning.
\newblock In {\em Proceedings of the IEEE conference on Computer Vision and
  Pattern Recognition}, pages 2001--2010, 2017.

\bibitem{saxe2011random}
Andrew~M Saxe, Pang~Wei Koh, Zhenghao Chen, Maneesh Bhand, Bipin Suresh, and
  Andrew~Y Ng.
\newblock On random weights and unsupervised feature learning.
\newblock In {\em Proceedings of the 28th International Conference on
  International Conference on Machine Learning}, pages 1089--1096, 2011.

\bibitem{sener2017active}
Ozan Sener and Silvio Savarese.
\newblock Active learning for convolutional neural networks: A core-set
  approach.
\newblock {\em arXiv preprint arXiv:1708.00489}, 2017.

\bibitem{sener2018active}
Ozan Sener and Silvio Savarese.
\newblock Active learning for convolutional neural networks: A core-set
  approach.
\newblock In {\em International Conference on Learning Representations}, 2018.

\bibitem{vgg}
Karen Simonyan and Andrew Zisserman.
\newblock Very deep convolutional networks for large-scale image recognition.
\newblock In {\em International Conference on Learning Representations}, May
  2015.

\bibitem{such2020generative}
Felipe~Petroski Such, Aditya Rawal, Joel Lehman, Kenneth Stanley, and Jeffrey
  Clune.
\newblock Generative teaching networks: Accelerating neural architecture search
  by learning to generate synthetic training data.
\newblock In {\em International Conference on Machine Learning}, pages
  9206--9216. PMLR, 2020.

\bibitem{sucholutsky2020secdd}
Ilia Sucholutsky and Matthias Schonlau.
\newblock Secdd: Efficient and secure method for remotely training neural
  networks.
\newblock {\em arXiv preprint arXiv:2009.09155}, 2020.

\bibitem{sucholutsky2021soft}
Ilia Sucholutsky and Matthias Schonlau.
\newblock Soft-label dataset distillation and text dataset distillation.
\newblock In {\em 2021 International Joint Conference on Neural Networks
  (IJCNN)}, pages 1--8. IEEE, 2021.

\bibitem{tian2020contrastive}
Yonglong Tian, Dilip Krishnan, and Phillip Isola.
\newblock Contrastive multiview coding.
\newblock In {\em European conference on computer vision}, pages 776--794.
  Springer, 2020.

\bibitem{toneva2018empirical}
Mariya Toneva, Alessandro Sordoni, Remi Tachet~des Combes, Adam Trischler,
  Yoshua Bengio, and Geoffrey~J Gordon.
\newblock An empirical study of example forgetting during deep neural network
  learning.
\newblock {\em arXiv preprint arXiv:1812.05159}, 2018.

\bibitem{ulyanov2016instance}
Dmitry Ulyanov, Andrea Vedaldi, and Victor Lempitsky.
\newblock Instance normalization: The missing ingredient for fast stylization.
\newblock {\em arXiv preprint arXiv:1607.08022}, 2016.

\bibitem{wang2022cafe}
Kai Wang, Bo Zhao, Xiangyu Peng, Zheng Zhu, Shuo Yang, Shuo Wang, Guan Huang,
  Hakan Bilen, Xinchao Wang, and Yang You.
\newblock Cafe: Learning to condense dataset by aligning features.
\newblock In {\em Proceedings of the IEEE/CVF Conference on Computer Vision and
  Pattern Recognition}, pages 12196--12205, 2022.

\bibitem{wang2018dataset}
Tongzhou Wang, Jun-Yan Zhu, Antonio Torralba, and Alexei~A Efros.
\newblock Dataset distillation.
\newblock {\em arXiv preprint arXiv:1811.10959}, 2018.

\bibitem{wei2015submodularity}
Kai Wei, Rishabh Iyer, and Jeff Bilmes.
\newblock Submodularity in data subset selection and active learning.
\newblock In {\em International Conference on Machine Learning}, pages
  1954--1963. PMLR, 2015.

\bibitem{welling2009herding}
Max Welling.
\newblock Herding dynamical weights to learn.
\newblock In {\em Proceedings of the 26th Annual International Conference on
  Machine Learning}, pages 1121--1128, 2009.

\bibitem{wolf2011facility}
Gert~W Wolf.
\newblock Facility location: concepts, models, algorithms and case studies.
  series: Contributions to management science.
\newblock {\em International Journal of Geographical Information Science},
  25(2):331--333, 2011.

\bibitem{xiao2017fashion}
Han Xiao, Kashif Rasul, and Roland Vollgraf.
\newblock Fashion-mnist: a novel image dataset for benchmarking machine
  learning algorithms.
\newblock {\em arXiv preprint arXiv:1708.07747}, 2017.

\bibitem{yoon2021online}
Jaehong Yoon, Divyam Madaan, Eunho Yang, and Sung~Ju Hwang.
\newblock Online coreset selection for rehearsal-based continual learning.
\newblock {\em arXiv preprint arXiv:2106.01085}, 2021.

\bibitem{zhao2021DSA}
Bo Zhao and Hakan Bilen.
\newblock Dataset condensation with differentiable siamese augmentation.
\newblock In {\em International Conference on Machine Learning}, pages
  12674--12685. PMLR, 2021.

\bibitem{zhao2021DM}
Bo Zhao and Hakan Bilen.
\newblock Dataset condensation with distribution matching.
\newblock {\em arXiv preprint arXiv:2110.04181}, 2021.

\bibitem{zhao2022synthesizing}
Bo Zhao and Hakan Bilen.
\newblock Synthesizing informative training samples with gan.
\newblock {\em arXiv preprint arXiv:2204.07513}, 2022.

\bibitem{zhao2021dataset}
Bo Zhao, Konda~Reddy Mopuri, and Hakan Bilen.
\newblock Dataset condensation with gradient matching.
\newblock In {\em Ninth International Conference on Learning Representations
  2021}, 2021.

\bibitem{NMI21}
H.-Y. Zhou, X. Chen, Y. Zhang, R. Luo, L. Wang, and Y. Yu.
\newblock Generalized radiograph representation learning via cross-supervision
  between images and free-text radiology reports.
\newblock {\em Nature Machine Intelligence}, 4:32–40, 2022.

\bibitem{zhou2022dataset}
Yongchao Zhou, Ehsan Nezhadarya, and Jimmy Ba.
\newblock Dataset distillation using neural feature regression.
\newblock {\em arXiv preprint arXiv:2206.00719}, 2022.

\bibitem{zhou2020distilled}
Yanlin Zhou, George Pu, Xiyao Ma, Xiaolin Li, and Dapeng Wu.
\newblock Distilled one-shot federated learning.
\newblock {\em arXiv preprint arXiv:2009.07999}, 2020.

\end{thebibliography}
}

\newpage

\setcounter{section}{0}
\section{Implementation Details}

\noindent \textbf{Random, Herding and Forgetting for ImageNet Subset.} Following other methods in Table~1 \yq{of the main paper}, we use the augmentation strategy of DC\cite{zhao2021dataset} in the evaluation of three baselines of ImageNet Subset. For \textbf{Random}, we randomly pick samples from the real set according to the experiment setting and evaluate the performance using the same model architecture and hyper-parameters. For \textbf{Herding}, we first pretrain the ConvNet using the entire ImageNet Subset for 40 epochs, where the accuracy reaches the peak performance, then use the ConvNet to extract features for all images. Finally, we calculate the average feature of all features from the same class and select \yq{its} closest neighbors among the features of the real images as the coreset for evaluation. Similar to DM\cite{zhao2021DM}, we use the L2 distance for closest neighbor calculation. For \textbf{Forgetting}, we first train the ConvNet with the entire ImageNet Subset and \yq{then} count the number of epoch-wise incorrect predictions for each real sample independently. Finally, we select the samples with the largest numbers \yq{of incorrect predictions} as the coreset. Since the performance of the model continues to increase as the training progresses, we empirically found that \textbf{Forgetting} performs \yq{much} worse than the other two baselines when the epoch number is set to a large number, possibly because ImageNet contains some mislabeled samples. Therefore, we \yq{reduce} the epoch number for ConvNet training \yq{to} 5 and randomly select the samples with the largest number\yq{s of incorrect epoch-wise predictions}.

\vspace{2mm}
\noindent \textbf{Other Details of Our Method.} 
For all the experiments in Table~1 \yq{of the main paper}, we \yq{use a} learning rate \yq{of} 0.2. Besides, we set an extra interval for the \textbf{Push} and \textbf{Train} model queue operations to reduce the training cost for condensation. For CIFAR-10/100, we set the interval as 30.

\section{Cross-architectural Experiments}


\yq{As a complement to Table~4, Sec.~5.5 of the main paper},
we further compare our method to DM with three other architectures. The \yq{full table is} shown in Table~\ref{table:cross_arch}. \yq{It can be observed that} our method outperforms DM in all 4$\times$4 different settings.

\begin{table}
\centering
\caption{Cross-architectural performance of our IDM method on CIFAR-10 with 10 Img/Cls. Our IDM achieves a significant improvement over DM. \yq{{\bf Bold}: DM with three other architectures that are not included in the main paper.}}
\label{table:cross_arch}
\resizebox{0.85\columnwidth}{!}{
\begin{tabular}{l|c|c|c|c|c} 
\toprule
                        & C$\backslash$T    & ConvNet  & AlexNet & VGG & ResNet  \\ 
\midrule
\multirow{4}{*}{DM}       & ConvNet & 48.9±0.6       & 38.8±0.5  & 42.1±0.4 & 41.2±1.1  \\
& {\bf AlexNet} & 34.4±0.3 & 28.8±1.1 & 31.6±0.6 & 31.4±0.3 \\
& {\bf VGG}     & 31.7±0.7 & 30.1±1.3 & 31.9±0.4 & 30.0±1.0 \\
& {\bf ResNet}  & 35.5±0.3 & 31.3±0.3 & 32.6±0.7 & 35.3±0.9 \\
\midrule
\multirow{4}{*}{Ours}     & ConvNet & 53.0±0.3 & 44.6±0.8 & 47.8±1.1 & 44.6±0.4        \\
                        & AlexNet & 44.8±0.5 & 41.4±1.4        & 43.1±0.6    & 41.0±0.1        \\
                        & VGG     & 41.2±0.4 & 37.4±0.3        & 41.7±0.4    &  38.8±0.8       \\ 
                        & ResNet  & 38.3±0.4 & 37.0±0.7        & 39.0±0.1    & 39.0±0.4        \\
\midrule
\end{tabular}
}
\end{table}

\section{Start with Pre-trained Models}

As we mentioned in Sec 4.2 of the main paper, our method pushes randomly initialized networks to the model queue, and the model fetched from the model queue is trained for $K$ iterations in each condensation step. However, when the model queue is applied to larger and more difficult datasets, sometimes we need a large $K$ and $N_{max}$ to make the model queue large enough to contain sufficiently trained models, i.e., the model trained for $K \times N_{max}$ iteration is well-performing enough to extract meaningful embeddings. This could be computationally intensive. 
To alleviate the problem, we propose some simple modifications to the model queue that can enable the model queue to start with pretrained networks, i.e., $P_\theta(t)$ where $t > 0$. 

The first modification for the problem is a small pre-trained model set, in which the models are trained for $t$ iteration, and the \textbf{Push} operation of the model queue pushes a randomly selected model from the pre-trained model set to the model queue. Besides, to preserve the diversity of the starting models for the model queue, we further apply some changes to the optimization scheme for each selected model. When pushing each pre-trained model to the model queue, we initialize the corresponding optimizer with a smaller random perturbation to the learning rate:
\begin{equation}
    lr^{*} = lr + Random(-0.1\times lr, 0.1\times lr),
\end{equation}
where $lr$ is the original learning rate, $Random(-0.1\times lr, 0.1 \times lr)$ generates a random float value between $-0.1\times lr$ and $0.1 \times lr$, and $lr*$ is the new learning rate for the corresponding optimizer. 
Moreover, we randomly select a subset of image classes and assign the subset to the selected model. In the training operation of the model queue, we only trained the model with the real image samples from the categories in the subset, thus the optimization of each pre-trained model can be diversified for better condensation.
The learning rate perturbation and class subset training can be viewed as some model augmentation techniques for model diversity preservation. We use the techniques in the experiments of ImageNet Subset to reduce training effort. 

\begin{figure*}[t]
  \centering
   \includegraphics[width=0.5\linewidth]{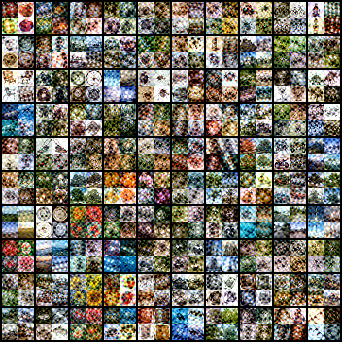}

  \caption{Visualization of synthetic set on CIFAR-100 with 1 image per class {\bf with} our partition and expansion augmentation.}

   \label{fig:c100_ipc1_aug}
\end{figure*}



\section{Visualization of Condensed Synthetic Sets}

\begin{figure*}[t]
  \centering
   \includegraphics[width=0.5\linewidth]{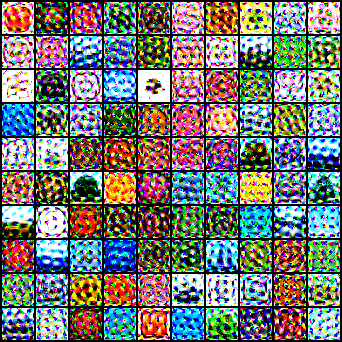}

  \caption{Visualization of synthetic set on CIFAR-100 with 1 image per class {\bf without} our partition and expansion augmentation.}

   \label{fig:c100_ipc1}
\end{figure*}

\begin{figure*}[t]
  \centering
   \includegraphics[width=0.5\linewidth]{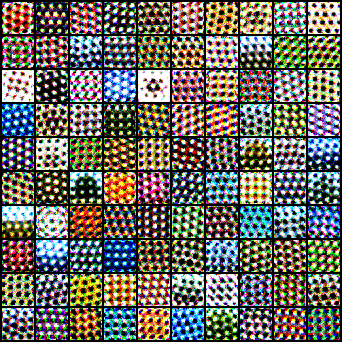}

  \caption{({\bf Original DM\cite{zhao2021DM} results}) Visualization of synthetic set on CIFAR-100 with 1 image per class {\bf without} partition and expansion augmentation.}

   \label{fig:dm_c100_ipc1}
\end{figure*}

\begin{figure*}[t]
  \centering
   \includegraphics[width=0.5\linewidth]{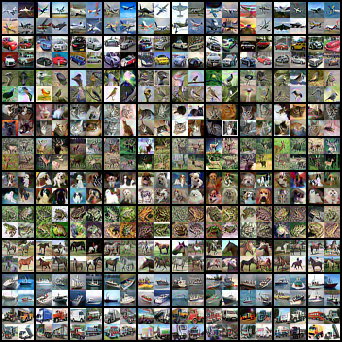}

  \caption{Visualization of synthetic set on CIFAR-10 with 10 image per class {\bf with} our partition and expansion augmentation.}

   \label{fig:c10_ipc10_aug}
\end{figure*}

\begin{figure*}[t]
  \centering
   \includegraphics[width=0.5\linewidth]{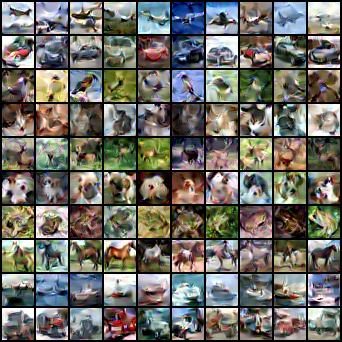}

  \caption{Visualization of synthetic set on CIFAR-10 with 10 image per class {\bf without} our partition and expansion augmentation.}

   \label{fig:c10_ipc10}
\end{figure*}

Fig.~\ref{fig:c100_ipc1_aug} and Fig.~\ref{fig:c100_ipc1} visualize the synthetic sets condensed by our method on CIFAR-100 with 1 image per class with and without the proposed partition and expansion augmentation, respectively. 
Interestingly, there are some repetitive textures in all images of Fig.~\ref{fig:c100_ipc1}, which might indicate lower utilization of pixels as fine-grained image details are discarded during forward propagation. 
For more information, we also visualize the results of original DM\cite{zhao2021DM} in Fig.~\ref{fig:dm_c100_ipc1}, whose textures share similar patterns to ours.
Fig.~\ref{fig:c10_ipc10_aug} and Fig.~\ref{fig:c10_ipc10} visualize the results of CIFAR-10 with 10 images per class with/without our augmentation respectively.

\end{document}